\documentclass[numsec,webpdf,modern,medium]{oup-authoring-template}
\usepackage{xurl}
\usepackage{array}
\usepackage{hyperref}
\usepackage{booktabs}

\graphicspath{{Fig/}}

\theoremstyle{thmstyleone}%
\theoremstyle{thmstyletwo}%
\theoremstyle{thmstylethree}%

\begin{document}

\journaltitle{Journal Title Here}
\DOI{DOI added during production}
\copyrightyear{YEAR}
\pubyear{YEAR}
\vol{XX}
\issue{x}
\access{Published: Date added during production}
\appnotes{Paper}

\firstpage{1}


\title[Neutral Prompts, Non-Neutral People]{Neutral Prompts, Non-Neutral People: Quantifying Gender and Skin-Tone Bias in Gemini Flash 2.5 Image and GPT Image 1.5}

\author[1,$\ast$]{Roberto Balestri\ORCID{0009-0000-5008-2911}}

\address[1]{\orgdiv{Department of the Arts}, \orgname{Università di Bologna}, \orgaddress{\country{Italy}}}

\corresp[$\ast$]{Corresponding author. \href{email:roberto.balestri2@unibo.it}{roberto.balestri2@unibo.it}}

\received{Date}{0}{Year}
\revised{Date}{0}{Year}
\accepted{Date}{0}{Year}



\abstract{
This study quantifies gender and skin-tone bias in two widely deployed commercial image generators—Gemini Flash 2.5 Image (NanoBanana) and GPT Image 1.5—to test the assumption that neutral prompts yield demographically neutral outputs.
We generated 3,200 photorealistic images using four semantically neutral prompts. The analysis employed a rigorous pipeline combining hybrid color normalization, facial landmark masking, and perceptually uniform skin tone quantification using the Monk (MST), PERLA, and Fitzpatrick scales.
Neutral prompts produced highly polarized defaults. Both models exhibited a strong ``default white'' bias ($>96$\% of outputs). However, they diverged sharply on gender: Gemini favored female-presenting subjects, while GPT favored male-presenting subjects with lighter skin tones.
This research provides a large-scale, comparative audit of state-of-the-art models using an illumination-aware colorimetric methodology, distinguishing aesthetic rendering from underlying pigmentation in synthetic imagery.
The study demonstrates that neutral prompts function as diagnostic probes rather than neutral instructions. It offers a robust framework for auditing algorithmic visual culture and challenges the sociolinguistic assumption that unmarked language results in inclusive representation.
}


\maketitle


\section{Introduction}

Generative image models have rapidly transitioned from experimental research artifacts to widely deployed commercial systems embedded in creative tools, productivity software, and online platforms. Systems capable of producing photorealistic images from short textual descriptions now influence advertising, education, entertainment, journalism, and everyday communication \citep{ali2024demographic, bianchi2023easily}. As a result, the representations produced by these models increasingly shape how users visualize abstract concepts such as ``a person,'' ``someone,'' or ``a human being'' \citep{ghosh2023person}. Despite this growing influence, the implicit demographic assumptions encoded within these systems remain insufficiently examined.

When an image generation model is prompted with a neutral description of a human, the resulting image is not merely a technical output; it is a normative statement about what constitutes a ``default'' person. These defaults matter. They influence who is seen, who is centered, and who is rendered invisible \citep{wolfe2022markedness, kay2015unequal}. In global contexts where users span diverse ethnicities, cultures, and identities, biased defaults can reinforce structural inequities and perpetuate historical patterns of exclusion \citep{frankenburg1993white}.

Previous research has demonstrated that algorithmic systems trained on large-scale datasets often inherit biases present in their source material \citep{radford2021learning, guilbeault2024online}. In computer vision, this has manifested as uneven performance across demographic groups in tasks such as face recognition, emotion detection, and age estimation \citep{buolamwini2018gender, grother2019face}. In generative systems, however, bias operates differently: rather than misclassifying individuals, models actively construct representations of people. This generative capacity introduces new ethical and methodological challenges \citep{friedrich2023fair, wang2023t2iat}. Instead of asking whether a system recognizes a face correctly, we must ask which faces it chooses to generate when given no explicit guidance.

A common assumption in both industry discourse and casual usage is that neutral prompts produce neutral results. Terms such as ``a person'' or ``an individual'' are often treated as demographically unmarked, implicitly inclusive descriptors. However, sociolinguistic research has long shown that \textit{neutral} language frequently masks dominant cultural norms \citep{sczesny2016can, spinelli2023neutral}. In many Western contexts, neutrality is implicitly aligned with whiteness, maleness, and youth. Whether contemporary AI image generators replicate, amplify, or diverge from these norms remains an open empirical question \citep{bianchi2023easily, ulloa2024representativeness}.

This study focuses on two commercially deployed, \textit{state-of-the-art} image generation systems: \textit{Gemini Flash 2.5 Image (NanoBanana)} and \textit{GPT Image 1.5}. Both models are marketed as general-purpose, photorealistic image generators and are accessible via commercial APIs. They are representative of systems that millions of users may encounter directly or indirectly through downstream applications. Importantly, these models are developed by different organizations, trained on different datasets, and tuned using distinct optimization and alignment strategies \citep{bansal2022well}. Comparing their outputs under identical conditions provides a rare opportunity to examine how different design choices influence demographic defaults. For the sake of brevity, the remainder of this paper will refer to these systems simply as \textit{NanoBanana} and \textit{GPT}, respectively.

The present work investigates two tightly coupled dimensions of representational bias: \textit{gender} and \textit{skin tone}. Gender is one of the most immediately legible human attributes and has been shown to exhibit strong skew in both recognition and generation tasks \citep{kay2015unequal, gleason2024perceptions}. Skin tone, meanwhile, is a critical axis of representation that intersects with race, ethnicity, and colorism \citep{monk2023scale, barrett2023skin}. Unlike coarse racial categories, skin tone can be measured continuously and mapped to established dermatological scales, enabling more nuanced quantitative analysis \citep{fitzpatrick1988validity, heldreth2024skin}.

Auditing skin tone in generated images poses significant technical challenges. Generative models frequently apply cinematic lighting, stylistic color grading, and aesthetic post-processing that alter pixel-level color values without changing perceived identity. Warm lighting, soft shadows, and makeup effects can all shift measured skin color independently of underlying pigmentation. Studies that rely on raw RGB values risk conflating lighting artifacts with genuine skin tone differences, leading to misleading conclusions \citep{thong2023beyond}.

To address this, we adopt a rigorous image analysis pipeline designed to separate illumination effects from pigmentation as much as possible. Our approach combines color space normalization, background-referenced white balance, precise facial landmark-based skin masking, and perceptually uniform color distance metrics \citep{hamrani2024ai}. By grounding skin tone measurement in \textit{CIELAB} space and mapping results to multiple dermatological scales, we aim to produce results that are both technically robust and interpretable across disciplines \citep{vansong2025new}.

Another challenge in auditing generative systems lies in scale and variance. Single prompts or small sample sizes are insufficient to characterize stochastic models whose outputs vary across runs. Accordingly, we conduct a large-scale generation study, producing 3,200 images across four semantically equivalent neutral prompts. This design allows us to examine not only aggregate tendencies, but also consistency across linguistic variations that users might reasonably expect to be interchangeable.

Importantly, this study does not attempt to infer the intentions of model developers, nor does it claim that observed biases are necessarily deliberate. Instead, we treat these systems as \textit{black boxes} and evaluate their observable behavior under controlled conditions \citep{raji2019actionable}. Our goal is descriptive and diagnostic: to quantify what these models do when asked to generate a ``person,'' not to speculate on why they do so beyond plausible technical explanations.

The contributions of this paper are fourfold:

\begin{enumerate}
    \item \textbf{Empirical evidence of demographic defaults:} We provide large-scale quantitative evidence that neutral prompts do not yield demographically neutral outputs in commercial image generators.
    \item \textbf{Comparative analysis across vendors:} By evaluating two leading systems under identical conditions, we show that demographic bias is not uniform across models and can manifest in opposing directions.
    \item \textbf{Methodological advances in skin tone auditing:} We introduce a hybrid normalization and masking pipeline that mitigates common pitfalls in skin tone measurement from generated imagery.
    \item \textbf{Intersectional insights:} We demonstrate systematic interactions between gender and skin tone in generated images, revealing aesthetic biases that would be obscured by single-axis analysis.
\end{enumerate}

The remainder of this paper is structured as follows. Section 2 contains previous works and research about the topic. Section 3 describes the experimental methodology, including prompt design, image generation, preprocessing, and analytical techniques. Section 4 presents the quantitative results, detailing gender distributions, racial classifications, and skin tone metrics. Section 5 discusses the implications of these findings in the context of representational harm, aesthetic bias, and model alignment. Section 6 outlines limitations and directions for future work, and Section 7 concludes with recommendations for auditing and deploying generative image systems responsibly.

\section{Related Work}

Research on demographic bias in machine vision, text-to-image generation, and skin tone measurement provides the conceptual and methodological foundation for this study. This section synthesizes four intersecting strands of work:
\begin{enumerate}[label=(\roman*)]
    \item demographic bias in facial analysis;
    \item skin tone as a fairness axis and measurement target;
    \item representational bias in online visual media and generative models;
    \item evidence that ostensibly neutral labels such as ``person'' encode whiteness and maleness as defaults.
\end{enumerate}

\subsection{Demographic Bias in Facial Analysis and Computer Vision}

Early work in algorithmic fairness established that commercial facial analysis systems exhibit large and systematic performance disparities across demographic groups. Buolamwini and Gebru's \textit{Gender Shades} audit demonstrated that gender classifiers from major vendors achieved near-perfect accuracy on lighter-skinned men while misclassifying darker-skinned women at error rates exceeding 30\%, highlighting an intersectional failure rooted in training data imbalance and evaluation practices \citep{buolamwini2018gender}. Subsequent studies confirmed that similar disparities persist across face recognition, age estimation, and emotion recognition pipelines, particularly under high-security thresholds.

Follow-up work on \textit{Actionable Auditing} showed that public, disaggregated evaluations can prompt meaningful changes: after the release of \textit{Gender Shades}, targeted vendors updated their systems and substantially reduced error gaps, especially for darker-skinned women \citep{raji2019actionable}. Large-scale evaluations such as the NIST \textit{Face Recognition Vendor Test (FRVT)} further revealed orders-of-magnitude differences in false positive rates across demographic groups, underscoring that bias is systemic rather than model-specific \citep{grother2019face}.

Collectively, this literature reframes facial analysis as a site where representational injustice and unequal error allocation are structurally embedded, motivating similar scrutiny of generative systems that synthesize faces rather than merely classify them.

\subsection{Skin Tone as a Fairness Axis and Measurement Target}

Because racial categories are socially contingent and inconsistently operationalized, several authors argue that skin tone provides a more technically meaningful axis for evaluating visual fairness. Barrett et al. review skin tone annotation practices in computer vision and document wide variation in scales, binning strategies, and uncertainty reporting; many studies conflate race and tone or rely on \textit{ad hoc} categorizations without accounting for annotator disagreement \citep{barrett2023skin}.

The \textit{Fitzpatrick Skin Type} (FST) scale---originally developed to characterize sun response in predominantly white populations \citep{fitzpatrick1988validity}---has since been widely repurposed for AI evaluation, despite well-documented limitations in representing darker skin tones. Empirical evidence indicates that FST categories are perceived as less inclusive than alternative skin tone scales, particularly by individuals with darker skin \citep{heldreth2024skin}. Moreover, the scale inadequately captures variation among individuals who differ not only in skin luminance, but also in hue \citep{thong2023beyond}.

In response, the \textit{Monk Skin Tone} (MST) scale was introduced to better span the global skin tone spectrum and decouple tone from racial typologies \citep{monk2023scale}. Recent studies adopting MST consistently find overrepresentation of the lightest categories and severe underrepresentation of darker tones across medical imagery, advertising, and device testing.

At the same time, controlled experiments demonstrate that lighting, camera characteristics, and background context can shift apparent skin tone by more than one MST unit, complicating naive reliance on raw RGB values. Colorimetric approaches using perceptually uniform spaces such as \textit{CIELAB}, combined with normalization procedures, offer more stable estimates but remain sensitive to contextual factors such as makeup and localized redness \citep{hamrani2024ai, vansong2025new}.

\subsection{Representational Bias in Online Visual Media and Image Search}

A parallel body of work examines how visual platforms represent people by default. Image search engines have been shown to systematically underrepresent women and minoritized groups even for ostensibly neutral queries. Kay et al. and subsequent studies find that searches for occupations such as ``doctor'' or ``engineer'' skew heavily male, while generic queries like ``person'' continue to privilege men and depict women in more objectified ways \citep{kay2015unequal, ulloa2024representativeness}.

Large-scale analyses of online imagery reinforce these findings. \textit{Perceptions in Pixels} \citep{gleason2024perceptions} shows that generic people-related queries cluster around light skin tones (modal MST category 2) and younger age groups, while Guilbeault et al. \citep{guilbeault2024online} demonstrates that gender bias is amplified more strongly in images than in accompanying text across platforms such as Google, Wikipedia, and IMDb. Studies of advertising, dermatology imagery, and clinical image banks similarly report dominance of light skin tones (Fitzpatrick I–III), with darker tones either absent or confined to pathology-focused contexts.

This broader visual ecosystem forms a substantial portion of the training data for contemporary generative models and provides important context for interpreting their default outputs.

\subsection{Bias and Stereotypes in Text-to-Image Generation}

With the widespread deployment of diffusion-based text-to-image models, audits have begun to document how social stereotypes surface in generated imagery. Bianchi et al. show that prompts for occupations and traits---even when phrased without demographic qualifiers---produce images that amplify existing gender and racial disparities, and that explicit counter-stereotypical prompting has limited effect \citep{bianchi2023easily}.

Several studies propose formal bias metrics for generative systems. Wang et al. introduce the \textit{Text-to-Image Association Test} (T2IAT), demonstrating that \textit{Stable Diffusion} reproduces well-documented human biases, including associations between lighter skin and positive valence \citep{wang2023t2iat}. \textit{ENTIGEN} evaluates whether ethical language interventions can diversify outputs, finding modest gains but persistent sensitivity to prompt wording \citep{bansal2022well}.

Domain-specific audits echo these results: generated surgeons, scientists, and professionals are disproportionately male and white relative to real-world demographics \citep{ali2024demographic}. While bias mitigation techniques such as \textit{Fair Diffusion} can shift visible proportions, they typically operate \textit{post hoc} and do not interrogate models' internal representations of a generic human \citep{friedrich2023fair}.

Recent audits of multimodal systems further demonstrate that biases manifest jointly across text and image generation. Balestri \citep{balestri2024examining, balestri2025gender} provides an in-depth analysis of \textit{ChatGPT-4o}'s multimodal behavior and \textit{Gemini 2.0 Flash}'s linguistic bias, showing that sexual and female-specific content is moderated more strictly than violent or male-specific content, and that similar asymmetries emerge in visual generation tasks. Notably, the study finds that violent imagery involving male subjects is more readily generated than comparable imagery involving female subjects, suggesting that gendered norms influence not only semantic associations but also visual permissibility thresholds.

\subsection{Default Categories and the Semantics of ``Person''}

A closely related line of work examines how multimodal models operationalize generic human categories. Wolfe and Caliskan show that \textit{CLIP} preferentially applies the unmarked label ``person'' to white and male faces, while explicitly marking race or gender for others, a pattern they interpret through linguistic theories of \textit{markedness} \citep{wolfe2022markedness}. In related work, they demonstrate that \textit{CLIP}-like models associate American identity with whiteness, and that \textit{CLIP}-guided generators can actively lighten the skin of Black initialization images when prompted with national identity.

Ghosh and Caliskan extend this analysis to \textit{Stable Diffusion}, showing that the prompt ``a person'' yields images most similar to light-skinned Western men and least similar to non-binary or non-Western identities \citep{ghosh2023person}. They further document sexualization and homogenization effects for women from non-Western regions. This work suggests that generic labels function not as neutral descriptors but as carriers of dominant norms.

\subsection{Language, Neutrality, and Gendered/Racialized Defaults}

Sociolinguistic research has long shown that ostensibly neutral language often encodes dominant categories as defaults. Experimental studies demonstrate that generic masculine forms bias listeners toward male referents even under explicit instructions to interpret them generically \citep{sczesny2016can, spinelli2023neutral}. Critical whiteness scholarship similarly describes whiteness as an unmarked norm against which other identities are rendered visible and particular \citep{frankenburg1993white}.

\subsection{Positioning of the Present Study}

While existing work documents bias in facial analysis, image search, and text-to-image generation, several gaps remain. Few studies examine strictly neutral prompts as a means of probing models' default representations, and fewer still compare multiple commercial generators under matched conditions. Moreover, quantitative skin tone analysis using illumination-aware colorimetry and multiple dermatological scales remains rare, as does intersectional analysis of gender and skin tone.

The present study addresses these gaps by auditing two commercial image generators under large-scale neutral prompting, estimating gender, race, and skin tone using a normalized pipeline, and situating the results within broader theories of \textit{markedness} and neutrality. In doing so, it provides a comparative, intersectional account of how contemporary image generators visualize ``a person'' in the absence of any demographic guidance.

\section{Methodology}

This study adopts a quantitative, model-agnostic auditing approach to examine demographic bias in commercial AI image generation systems. The methodological design is driven by a central objective: to isolate and measure the implicit demographic defaults encoded in image generation models when prompted with language that is semantically neutral and demographically unspecified.

Achieving this requires careful control over prompt formulation, large-scale sampling to account for stochastic variation, and a robust image analysis pipeline capable of disentangling illumination effects from intrinsic skin pigmentation. Rather than relying on subjective visual inspection or coarse categorical labeling, we combine automated face analysis with perceptually grounded colorimetric techniques. This allows demographic tendencies to be measured consistently across thousands of generated images, while minimizing confounding factors introduced by stylistic rendering choices common in generative imagery.

\subsection{Model Selection and Scope}

This analysis examines two commercially deployed, \textit{state-of-the-art} image generation models: \textit{NanoBanana}, developed by Google and accessed via Google Cloud APIs, and \textit{GPT}, developed by OpenAI and accessed via the Azure API. Both models were evaluated using their most recent publicly available versions as of 28 January 2026. These models were selected because they represent widely available, general-purpose systems explicitly designed for photorealistic image synthesis and are likely to influence a broad range of downstream applications.

Importantly, the study treats both models as \textit{black boxes}. No assumptions are made regarding their internal architectures, training data composition, or fine-tuning procedures beyond publicly available documentation. This choice reflects the reality faced by most users, researchers, and policymakers, who interact with such systems and have limited visibility into their internal mechanisms. By focusing on observable outputs, the methodology remains applicable even as model internals evolve or remain proprietary.

The scope of the study is intentionally narrow in one respect and broad in another. It is narrow in that it examines only neutral prompt scenarios, without explicit demographic modifiers or stylistic constraints. It is broad in that it evaluates a large number of generations across multiple prompt phrasings, enabling statistically meaningful comparisons both within and between models.

\subsection{Prompt Design and Neutrality Constraints}

Prompt design plays a critical role in bias auditing. Subtle linguistic cues can activate cultural stereotypes or demographic expectations embedded in training data. To minimize such effects, prompts were deliberately constrained to short, semantically neutral descriptors referring to a human subject without specifying gender, ethnicity, age, profession, or context.

Four prompts were selected: ``a human being, photorealistic,'' ``a person, photorealistic,'' ``an individual, photorealistic,'' and ``someone, photorealistic.'' These phrases were chosen because they are commonly perceived as interchangeable in everyday language and lack explicit demographic markers. The inclusion of the term ``photorealistic'' serves to reduce stylistic abstraction and encourage outputs resembling real human faces rather than illustrations or stylized characters.

Each prompt was executed independently and repeatedly for each model. The repetition is essential because generative image models are stochastic systems: identical prompts can yield substantially different outputs across runs. By sampling each prompt 400 times per model, the study captures the underlying probability distribution of generated representations rather than isolated examples. 

Crucially, no attempt was made to balance or correct prompts to encourage diversity. This is a deliberate methodological choice. The goal is not to test whether models can generate diverse images when instructed to do so, but rather to observe what they generate by default when no such instruction is provided.

\subsection{Image Generation Conditions}

All images were generated using default API parameters provided by each platform at the time of data collection. Sampling strategy and internal randomness were left unchanged. Only resolution was fixed at 1024$\times$1024 pixels. No random seeds were fixed, and no post-generation filtering or selection was applied. This design choice reflects typical user behavior and avoids introducing artificial constraints that could bias the output distribution.

By maintaining parity in generation conditions across models, the methodology ensures that observed differences are attributable to model behavior rather than experimental artifacts. The resulting dataset consists of 3,200 images in total, evenly split between models and prompts.

\subsection{Preprocessing and Color Normalization}

A central methodological challenge in skin tone analysis lies in separating intrinsic pigmentation from extrinsic lighting effects. AI-generated images frequently employ warm color grading, soft shadows, and cinematic illumination that alter pixel values without changing perceived identity. Raw color measurements taken directly from such images risk systematically mischaracterizing skin tone.

To address this, we implement a hybrid color normalization pipeline designed to reduce illumination bias while preserving natural chromatic relationships. The pipeline operates in the \textit{CIELAB} color space, which is widely used in color science due to its perceptual uniformity.

The first stage applies \textit{contrast-limited adaptive histogram equalization} (CLAHE) to the lightness channel ($L^*$). This enhances local contrast in a controlled manner, mitigating extreme lighting variations without collapsing global tonal differences. However, \textit{CLAHE} alone can introduce artifacts or exaggerate texture, particularly in facial regions. To counteract this, the processed image is blended with the original image using equal weighting, producing a balanced representation that benefits from contrast enhancement without losing fidelity.

The final and most critical step is background-referenced white balance. Traditional white balance algorithms often assume that the average color of an image should be neutral, an assumption that fails when the subject's skin occupies a large portion of the frame. Instead, we estimate the illuminant using only background pixels. Facial landmarks are used to define and exclude the facial region, and the brightest background pixels are treated as a proxy for the scene's white point. Channel-wise correction is then applied uniformly across the image.

This approach ensures that warm-toned skin is not incorrectly ``neutralized'' while still compensating for global color casts. Visual inspection confirms that the majority of the resulting images appear more consistent in lighting without appearing artificially desaturated or altered.

\subsection{Face Detection and Alignment}

Following normalization, faces are detected using a deep learning-based face detection pipeline. Only images containing a clearly detectable face are retained for further analysis. In cases where multiple faces are present, only the largest face is processed, under the assumption that it represents the primary subject of the image. Due to the high fidelity enforced by the 'photorealistic' prompt constraint, the face detector successfully identified a primary face in all 3,200 generated samples.

Detected faces are aligned to a canonical orientation to reduce variance introduced by head pose. Alignment is particularly important for subsequent landmark-based masking, as it ensures that corresponding facial regions are sampled consistently across images.

\subsection{Landmark-Based Skin Region Extraction}

Accurate skin tone measurement depends critically on sampling the correct pixels. Many facial regions commonly included in naïve masks---such as lips, eyes, nostrils, or eyebrows---do not represent skin pigmentation and can significantly skew color measurements. Similarly, shadows near the eye sockets or specular highlights on the nose can introduce noise.

\begin{figure}[t]
    \centering
    \includegraphics[width=0.6\linewidth]{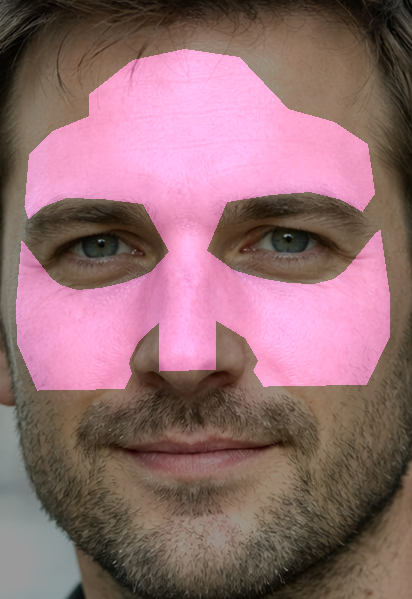}
    \caption{Visual demonstration of the landmark-based skin masking pipeline. The pink overlay indicates the valid pixel region retained for colorimetric analysis. Note the precise exclusion of non-skin features (eyes, eyebrows, nostrils, lips) and facial hair regions to prevent color contamination.}
    \label{fig:skin_mask}
    
    \vspace{5pt} 
    \small
    \textit{Alt text: A close-up image of a face with a solid pink digital mask overlaid on the skin. The mask covers the forehead, cheeks, nose, and chin but explicitly cuts out the eyes, eyebrows, nostrils, and lips, demonstrating how these features are excluded from skin tone analysis.}
\end{figure}

To address this, we employ a precise masking strategy based on dense facial landmarks. A convex hull is constructed from landmarks corresponding to broad skin-covered regions of the face, including the cheeks, jawline, and nasal bridge. From this base mask, regions corresponding to eyes, eyebrows, nostrils, and the mouth are explicitly removed using subtractive geometry. Figure \ref{fig:skin_mask} illustrates the final output of this masking procedure, highlighting the selective exclusion of high-noise facial features.

Standard facial landmark sets often terminate below the hairline, omitting much of the forehead. To compensate, we implement a tapered forehead expansion that extrapolates upward from the highest detected landmarks, scaled proportionally to facial height. This captures additional skin area while avoiding intrusion into hair regions.

The result is a high-confidence skin mask that prioritizes true skin surface while minimizing contamination from non-skin features, facial hair, or shadowed cavities.

\subsection{Skin Tone Quantification}

Skin tone quantification is performed exclusively within the masked skin region and in \textit{CIELAB} color space. Rather than relying on a single average color, which can be sensitive to highlights and shadows, we adopt a clustering-based approach inspired by dermatological image analysis methods.

Masked pixels are clustered using \textit{k-means} clustering, with four clusters selected to balance expressiveness and stability. Clusters are then ranked by pixel count, and the largest clusters are accumulated until they account for at least 36\% of all masked pixels. The centroids of these clusters are combined using a weighted average to produce a representative skin color for the image.

This method emphasizes dominant pigmentation while reducing the influence of outliers such as glare or deep shadow. The resulting color vector is then mapped to multiple established skin tone scales using perceptually uniform color distance metrics.

\subsection{Mapping to Dermatological Scales}
To ensure interpretability and cross-study comparability, measured skin tones are mapped to three dermatological scales: the Monk Skin Tone scale, the PERLA scale, and the Fitzpatrick Skin Type classification. Mapping is performed using Euclidean distance in CIELAB space, ensuring that assignments reflect perceived color similarity rather than arbitrary numeric thresholds.
Using multiple scales allows the analysis to be interpreted across different cultural and clinical contexts and reduces reliance on any single categorization scheme.
\subsection{Demographic Attribute Estimation}
In addition to skin tone, demographic attributes such as perceived gender, race, age, and dominant facial expression are estimated using a pretrained facial analysis framework. These attributes are not treated as ground truth but as consistent proxies that enable aggregation and intersectional analysis across the dataset.
Gender is treated as a binary classification reflecting the model’s perceived presentation rather than biological sex or self-identified gender. This framing aligns with the study’s focus on representation rather than identity.

\subsection{Statistical Aggregation}
Finally, results are aggregated across multiple dimensions, including model, prompt, and predicted gender. Summary statistics, distribution frequencies, and comparative means are computed to characterize both central tendencies and variability.
This multi-level aggregation enables the identification of systematic biases that persist across prompts, as well as differences that emerge between models or demographic groups.

\subsection{Implementation Details}

The entire auditing pipeline was implemented in Python 3.11. Face detection and demographic attribute estimation (age, race, and gender) were performed using the \textit{DeepFace}\footnote{\url{https://github.com/serengil/deepface}} library, utilizing \textit{RetinaFace} as the backend detector.

For precise skin region segmentation, we employed \textit{Google MediaPipe Face Mesh}\footnote{\url{https://ai.google.dev/edge/mediapipe/solutions/vision/face_landmarker}}, which provides a dense 468-point 3D face landmark model, allowing for the accurate exclusion of non-skin features such as eyes and lips. Colorimetric analysis and space conversions were handled using \textit{scikit-image}. All color distance calculations, including the mapping to dermatological scales, were computed using the CIEDE2000 ($\Delta E_{00}$) formula to ensure perceptual uniformity.

\section{Results}

This chapter presents the empirical findings of the study, organized around four interrelated dimensions:
\begin{enumerate}[label=(\roman*)]
    \item demographic defaults in neutral image generation, encompassing gender, race, and age distributions;
    \item skin tone distributions across multiple dermatological scales;
    \item systematic differences between models;
    \item intersectional effects between predicted gender and skin tone.
\end{enumerate}

All results are derived from the full dataset of 3,200 images generated using neutral prompts and processed using the hybrid normalization and skin analysis pipeline described in Chapter 3.

Results are reported separately for \textit{NanoBanana} and \textit{GPT} wherever this distinction is analytically meaningful. Importantly, \textit{Fitzpatrick Skin Type} (FST) is treated throughout as an ordinal categorical variable and is therefore analyzed using distributions, medians, and contingency-based statistics rather than means or standard deviations.

\subsection{Overall Demographic Distributions}

\subsubsection{Predicted Gender Distribution}

Across the full dataset, neutral prompts did not yield a balanced gender distribution. Of the 3,200 generated images, 1,969 (61.5\%) were classified as women and 1,231 (38.5\%) as men. However, as shown in Table \ref{table:gender_by_model}, this aggregate imbalance conceals extreme and opposing biases at the level of individual models.

Rather than converging toward parity, the two systems adopt opposite gender defaults. \textit{NanoBanana} overwhelmingly generates women, whereas \textit{GPT} overwhelmingly generates men. This polarity persists across all prompt formulations and demonstrates that neutrality in language does not translate into neutrality in generated representation.

A Chi-Square test of independence confirms that the gender distribution is significantly dependent on the model choice ($\chi^2 = 1395.19, p < 0.001$).

Aggregate results obscure substantial prompt-level variation that differs sharply between models. Table \ref{tab:gender_by_prompt} reports predicted gender distributions for each prompt, disaggregated by model.

\begin{figure*}[t]
    \centering
    \includegraphics[width=0.8\textwidth]{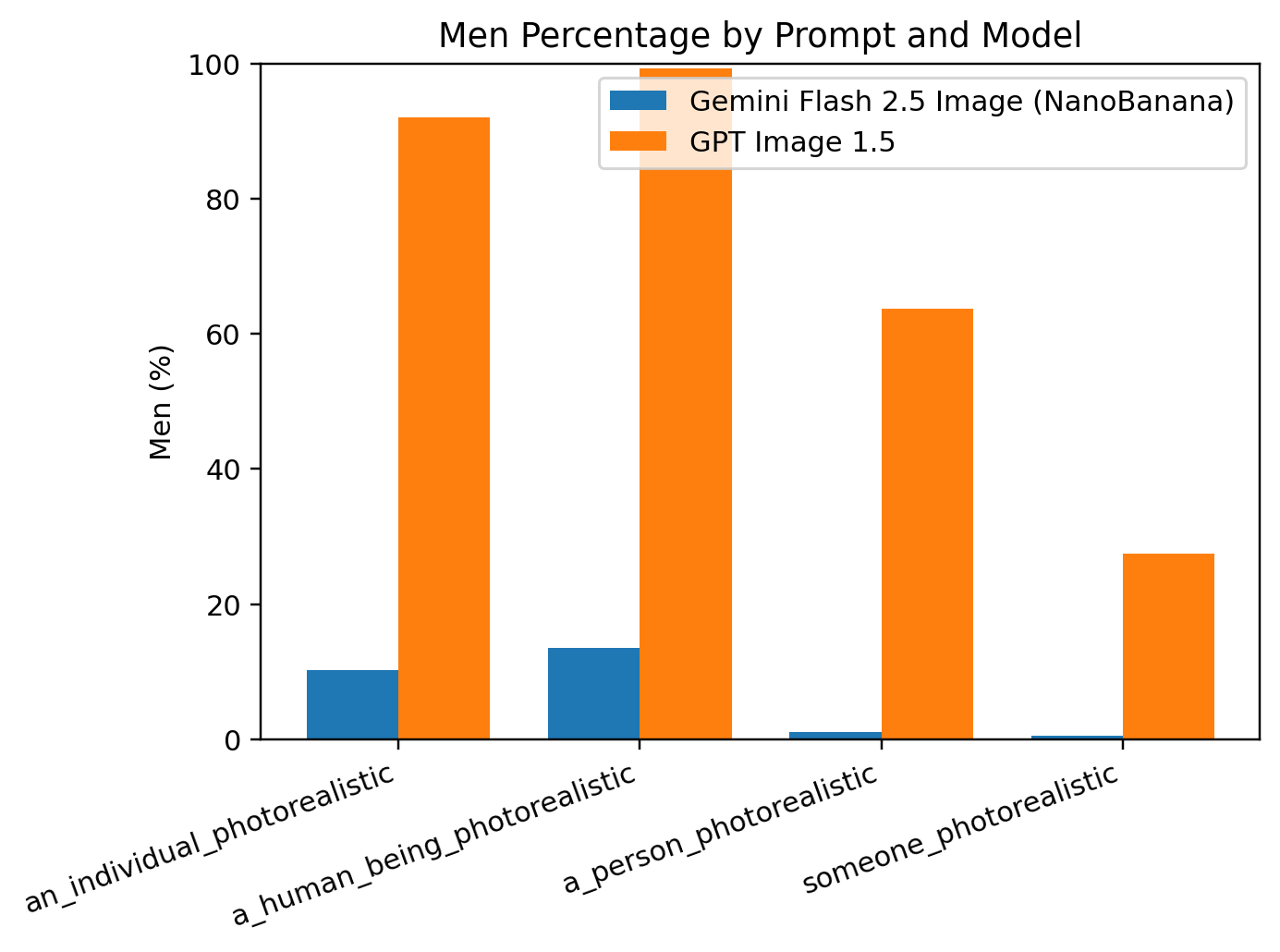}
    \caption{Percentage of generated subjects classified as men across four neutral prompts. The chart highlights the prompt-dependent variability in \textit{GPT} (orange) versus the consistently female-skewed output of \textit{NanoBanana} (blue). Note the sharp inversion for \textit{GPT} on the prompt ``someone,'' which flips from a male majority to a female majority.}
    \label{fig:gender_prompt_breakdown}

    \vspace{5pt}
    \small
    \textit{Alt text: A bar chart comparing the percentage of male subjects generated by GPT and NanoBanana across four neutral prompts. GPT shows high male percentages (over 60\%) for ``a human being,'' ``a person,'' and ``an individual,'' but drops sharply to a minority for ``someone.'' NanoBanana consistently shows very low male percentages (under 15\%) across all prompts.}
\end{figure*}

The same prompt produces diametrically opposite outcomes depending on the model, as visualized in Figure \ref{fig:gender_prompt_breakdown}. For example, ``a human being'' yields an almost exclusively male population under \textit{GPT}, but a strongly female population under \textit{NanoBanana}.

\textit{GPT} exhibits a particularly striking lexical sensitivity. While prompts such as ``a human being'' and ``an individual'' generate over 90\% male subjects, the prompt ``someone'' triggers a dramatic inversion of this default, shifting the distribution to 72.5\% female-presenting subjects. This anomaly suggests that while the model's broad default is male, specific lexical tokens such as ``someone'' carry distinct, female-coded associations within its latent space. In contrast, \textit{NanoBanana} remains consistently female-skewed across all formulations, with semantic variations having negligible impact on the demographic outcome.

These results demonstrate that prompt wording does not function independently. Instead, prompts act as triggers that reveal each model's latent demographic defaults, rather than neutralizing them.

\subsubsection{Predicted Racial Categories}

Predicted racial classifications show an even stronger collapse toward a single dominant category. As detailed in Table \ref{tab:race_distribution}, both models exhibit a profound lack of diversity when prompted with neutral descriptors.

Across both models, more than 96\% of generated individuals were classified as white. Representation of Black, Asian, and Middle Eastern individuals was minimal and, in some cases, entirely absent for a given model. These patterns were stable across prompt variants, suggesting that they reflect model-level priors rather than prompt-specific linguistic cues.

Despite the overwhelming dominance of white subjects in both models, the difference in racial distribution between NanoBanana and GPT is statistically significant ($\chi^2 = 94.07, p < 0.001$).

\subsubsection{Predicted Age Distribution}

In addition to gender and race, the study estimated the apparent age of generated subjects. Table \ref{tab:age_distribution} presents the mean predicted age disaggregated by prompt and model.

Consistent with the findings on gender and skin tone, the models exhibit distinct age defaults. \textit{NanoBanana} centers its representation on early adulthood, with an aggregate mean age of 26.5 years. By contrast, \textit{GPT} generates subjects that are, on average, over 5 years older (31.7 years).

While \textit{GPT} displays a slightly broader age range, both models exhibit a strong ``young adult'' bias. The default ``person'' is effectively conceptualized as an individual between 20 and 35 years old; neither children nor elderly individuals appear with significant frequency in the neutral generation set.

Consistent with the findings on gender and skin tone, the models exhibit distinct age defaults. \textit{NanoBanana} centers its representation on early adulthood, with an aggregate mean age of 26.5 years. By contrast, \textit{GPT} generates subjects that are, on average, over 5 years older (31.7 years). An independent samples t-test confirms that this difference of approximately 5.2 years is statistically significant ($t = -42.35, p < 0.001$).


\begin{table*}[b] 
    \centering
    \renewcommand{\arraystretch}{1.3} 

    \caption{Predicted Gender Distribution by Model}
    \label{table:gender_by_model}
    \begin{tabular*}{\textwidth}{@{\extracolsep{\fill}}lccc}
    \hline
    \textbf{Model} & \textbf{Total Images} & \textbf{Women} & \textbf{Men} \\ \hline
    \textit{NanoBanana} & 1,600 & 1,499 (93.7\%) & 101 (6.3\%) \\
    \textit{GPT} & 1,600 & 470 (29.4\%) & 1,130 (70.6\%) \\ \hline
    \textbf{Combined} & \textbf{3,200} & \textbf{1,969 (61.5\%)} & \textbf{1,231 (38.5\%)} \\ \hline
    \end{tabular*}

    \vspace{1.2cm}

    \caption{Predicted Gender Distribution by Prompt and Model (\%)}
    \label{tab:gender_by_prompt}
    \begin{tabular*}{\textwidth}{@{\extracolsep{\fill}}lcccc}
    \toprule
    \textbf{Prompt} & \multicolumn{2}{c}{\textbf{GPT}} & \multicolumn{2}{c}{\textbf{NanoBanana}} \\ 
    & Men (\%) & Women (\%) & Men (\%) & Women (\%) \\ \midrule
    a human being  & 99.2 & 0.8  & 13.5 & 86.5 \\
    a person       & 63.7 & 36.2 & 1.0  & 99.0 \\
    an individual  & 92.0 & 8.0  & 10.2 & 89.8 \\
    someone        & 27.5 & 72.5 & 0.5  & 99.5 \\ \bottomrule
    \end{tabular*}

    \vspace{1cm}

    \caption{Predicted Race Distribution across Models}
    \label{tab:race_distribution}
    \begin{tabular*}{\textwidth}{@{\extracolsep{\fill}}lccc}
    \toprule
    \textbf{Race Category} & \textbf{NanoBanana (n=1600)} & \textbf{GPT (n=1600)} & \textbf{Combined (n=3200)} \\ \midrule
    White            & 1,535 (95.9\%) & 1,550 (96.9\%) & 3,085 (96.4\%) \\
    Latino Hispanic  & 42 (2.6\%)     & 6 (0.4\%)      & 48 (1.5\%) \\
    Middle Eastern   & 23 (1.4\%)     & 0 (0.0\%)      & 23 (0.7\%) \\
    Black            & 0 (0.0\%)      & 38 (2.4\%)     & 38 (1.2\%) \\
    Asian            & 0 (0.0\%)      & 6 (0.4\%)      & 6 (0.2\%) \\ \bottomrule
    \end{tabular*}

    \vspace{1cm}

    \caption{Mean Predicted Age by Prompt and Model (Years)}
    \label{tab:age_distribution}
    \begin{tabular*}{\textwidth}{@{\extracolsep{\fill}}lcc}
    \toprule
    \textbf{Prompt} & \textbf{GPT Mean Age} & \textbf{NanoBanana Mean Age} \\ \midrule
    a human being  & 33.7 & 24.1 \\
    a person       & 31.2 & 28.1 \\
    an individual  & 32.9 & 25.3 \\
    someone        & 29.0 & 28.5 \\ \midrule
    \textbf{Overall Mean} & \textbf{31.7} & \textbf{26.5} \\ \bottomrule
    \end{tabular*}

\end{table*}


\mbox{} 
\clearpage

\begin{figure*}[t!]
    \centering
    \includegraphics[width=0.8\textwidth]{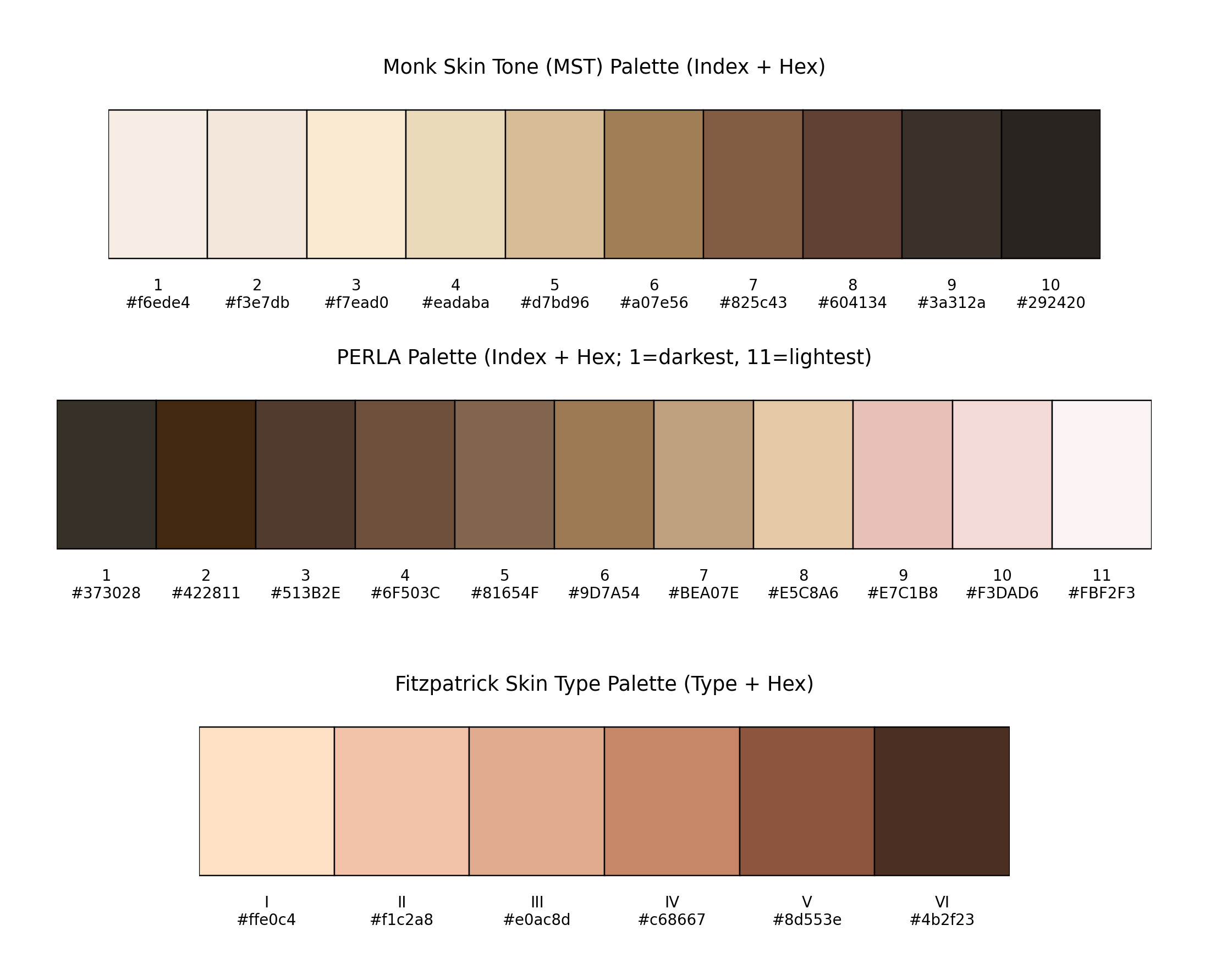}
    \caption{Reference color palettes for the three dermatological scales used in this study: Monk Skin Tone (MST), PERLA, and Fitzpatrick Skin Type (FST). Each scale provides a set of reference values against which generated skin tones are mapped using Euclidean distance in CIELAB space.}
    \label{fig:scales_palette}

    \vspace{5pt}
    \small
    \textit{Alt text: A composite image displaying the reference color swatches for three dermatological scales. The Monk Skin Tone (MST) scale shows a gradient of 10 tones from light to dark. The PERLA scale shows its specific color references. The Fitzpatrick Skin Type (FST) scale displays six distinct skin tone categories ranging from Type I (Pale White) to Type VI (Dark Brown/Black).}
\end{figure*}

\subsection{Skin Tone Analysis by Prompt and Model}
Skin tone was quantified using the \textit{Monk Skin Tone} (MST) and \textit{PERLA} scales, treated as quasi-continuous perceptual measures derived from \textit{CIELAB} color distances.

\subsubsection{Monk Skin Tone (MST)}

Table \ref{tab:mst_results} presents the mean MST values. Across all prompts, \textit{NanoBanana} produces systematically darker skin tones than \textit{GPT}. The difference is especially pronounced for ``a person'', where Gemini's mean MST exceeds GPT's by more than 1.6 points.

\textit{GPT} shows limited sensitivity to prompt wording in terms of skin tone, whereas \textit{NanoBanana} exhibits substantial prompt-driven variation.

\subsubsection{PERLA Skin Tone Scale}

\begin{figure*}[p] 
    \centering
    
    \includegraphics[width=0.7\textwidth]{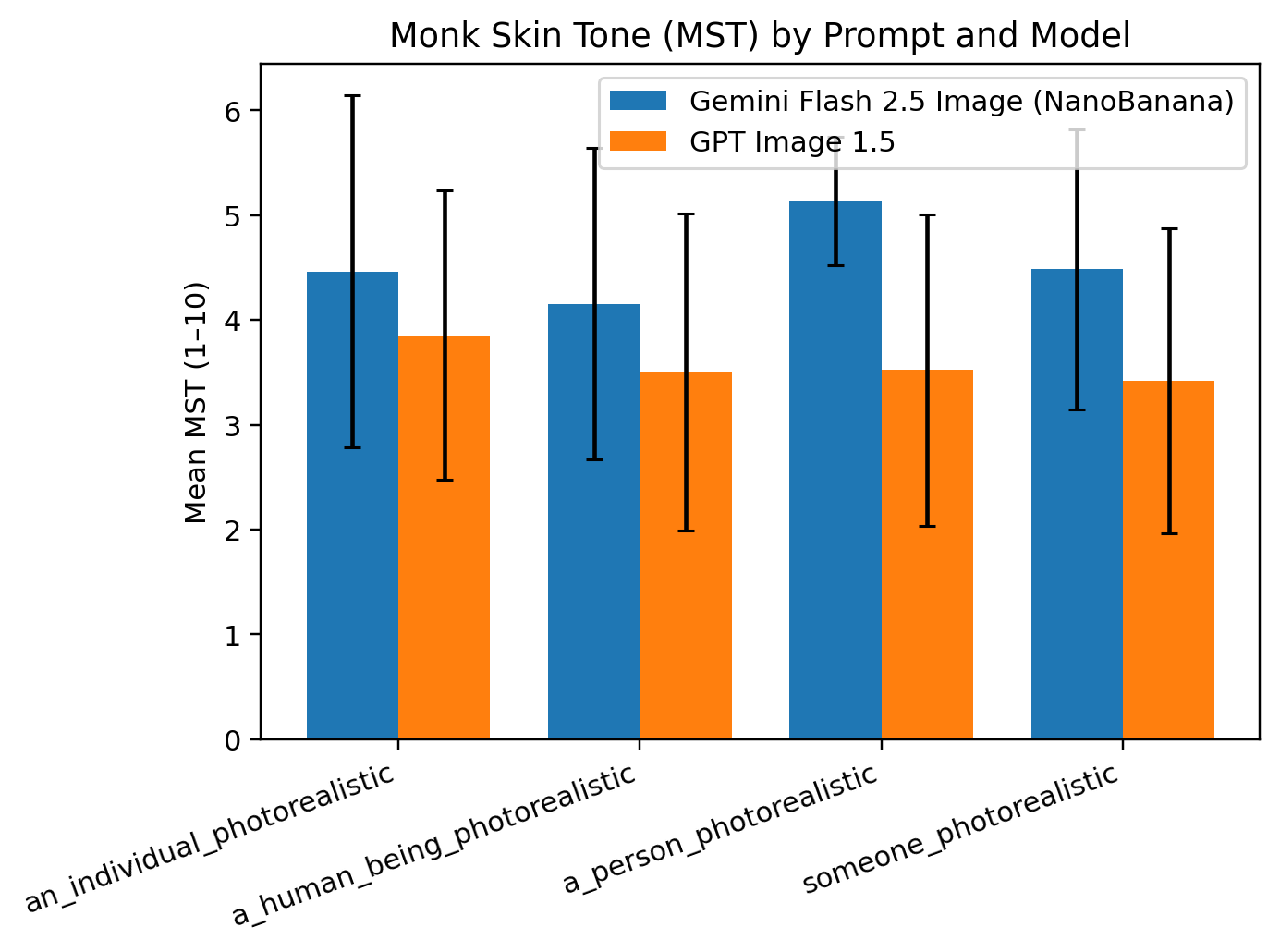}
    \captionof{figure}{Mean Monk Skin Tone (MST) scores by prompt and model (Scale 1--10, where 10=Darkest). Error bars represent standard deviation.}
    \label{fig:mst_results}
    \vspace{0.2cm}
    \small \textit{Alt text: A bar chart comparing mean Monk Skin Tone (MST) scores. NanoBanana (blue) shows higher values (4.2--5.1) with large error bars, while GPT (orange) shows lower values clustered around 3.5.}
    
    \vspace{1.5cm} 

    \includegraphics[width=0.7\textwidth]{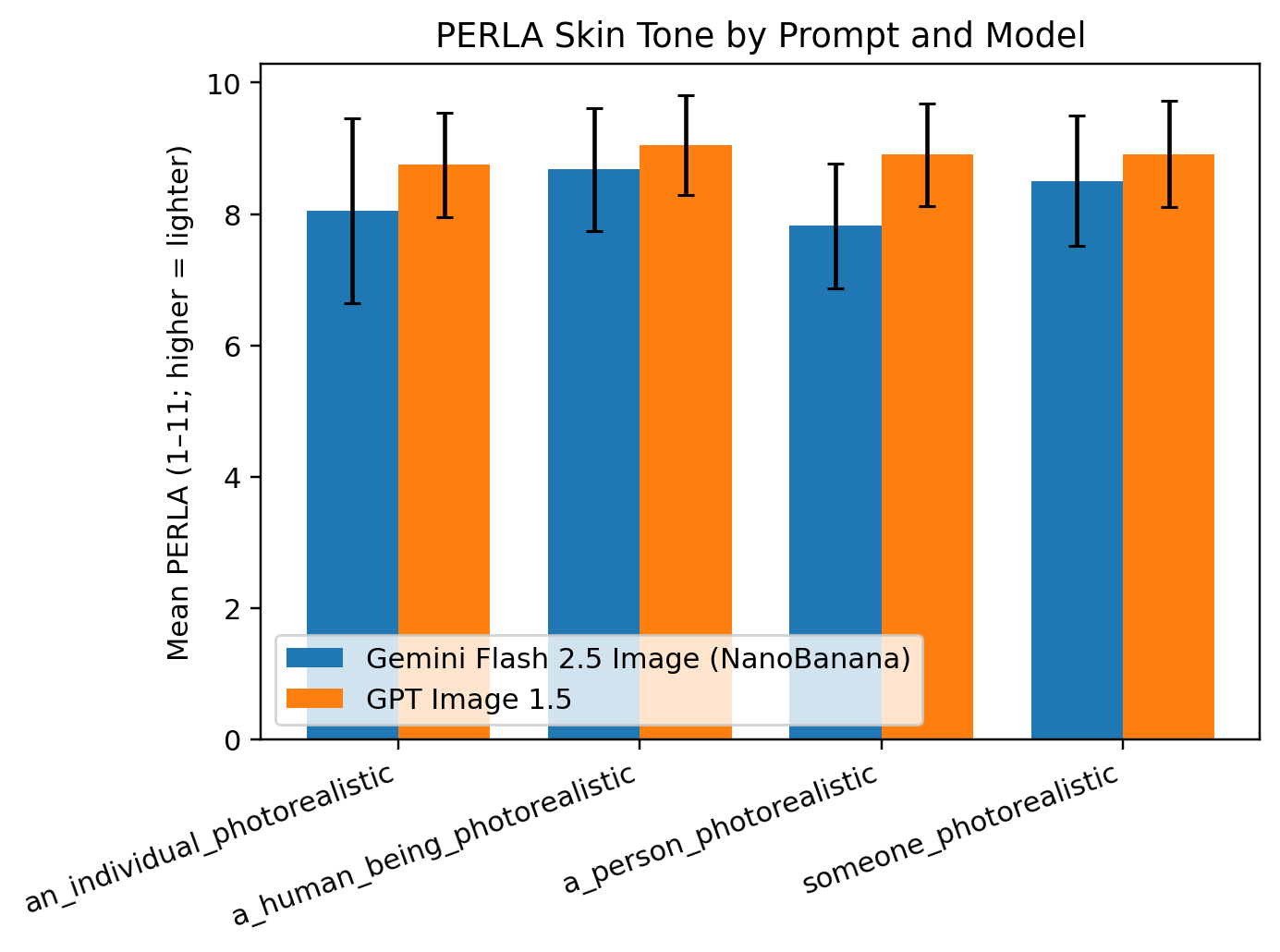}
    \captionof{figure}{Mean PERLA Skin Tone scores by prompt and model (Scale 1--11, where higher values = lighter skin). Error bars represent standard deviation.}
    \label{fig:perla_results}
    \vspace{0.2cm}
    \small \textit{Alt text: A bar chart of mean PERLA scores. GPT (orange) consistently scores high (8.8–9.0). NanoBanana (blue) scores lower and exhibits greater variability, notably for the prompt ``a person.''}

\end{figure*}

PERLA results mirror the findings from the MST analysis, as shown in Table \ref{tab:perla_results} and Figure \ref{fig:perla_results}. Lower PERLA values (which correspond to darker skin tones on this scale) are consistently observed in the \textit{NanoBanana} outputs compared to \textit{GPT}.

A Mann-Whitney U test confirms that this shift is statistically significant ($U=9.42 \times 10^5, p < 0.001$). As with the Monk scale, \textit{NanoBanana} exhibits considerably larger standard deviations across all prompts, suggesting a wider (though still biased) range of generated skin tones. \textit{GPT}, by contrast, remains tightly clustered near the top of the scale (PERLA 8.5--9.0), indicating a high degree of uniformity in generating light-skinned subjects regardless of the prompt used.

For the prompt ``a person,'' \textit{NanoBanana} drops to a mean PERLA score of 7.82, whereas \textit{GPT} remains at 8.90, further illustrating the prompt-dependent variability unique to the Google model.

\clearpage 


\begin{figure*}[t!]
    \centering
    \includegraphics[width=1\textwidth]{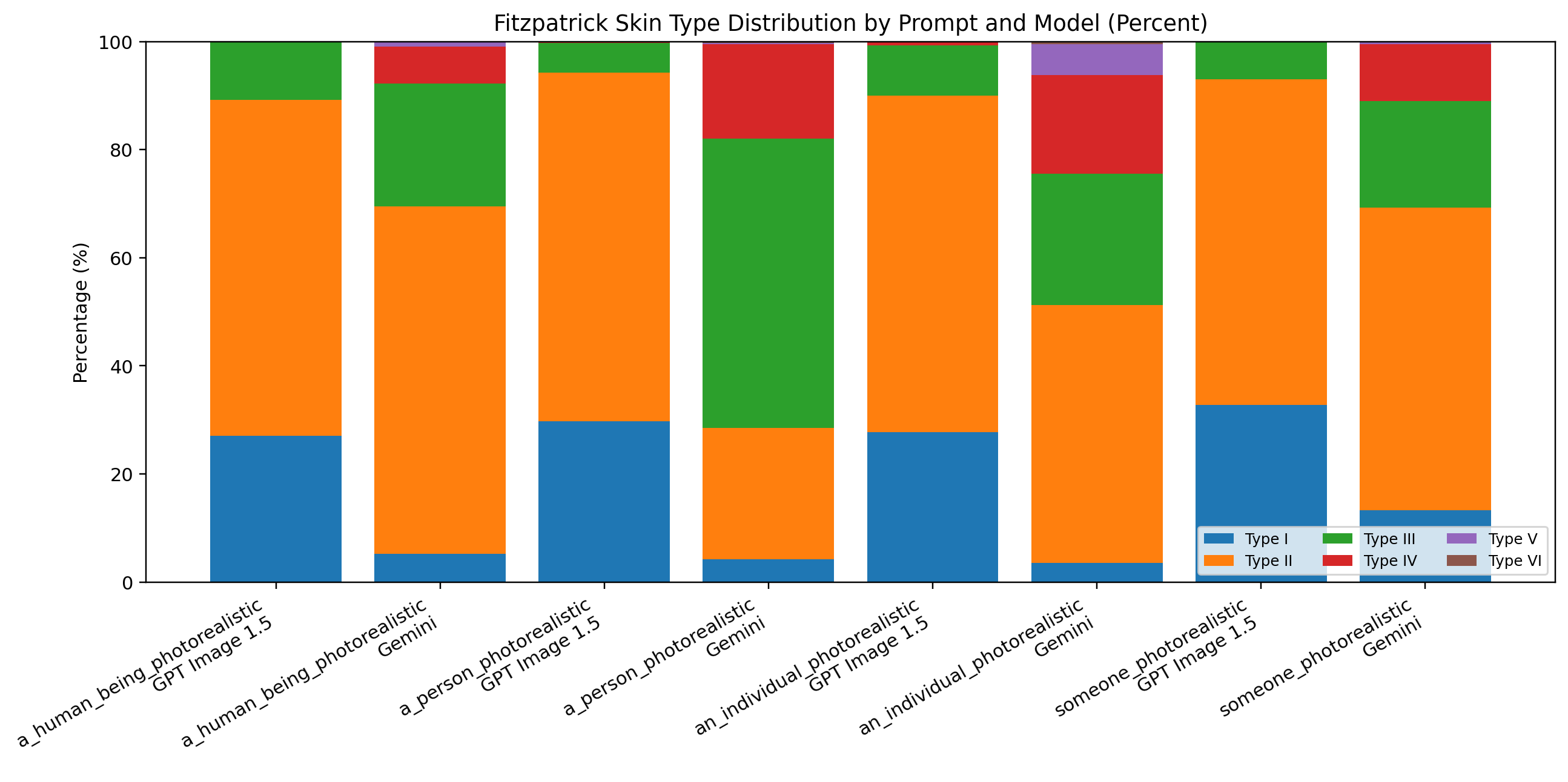}
    \caption{Stacked bar chart of Fitzpatrick Skin Type (FST) distribution by prompt and model. \textit{GPT} (left bars) is dominated by Types I--II (Blue/Orange), indicating a strong bias toward pale and fair skin. \textit{NanoBanana} (right bars) shows a broader distribution, extending significantly into Types III (Green) and IV (Red).}
    \label{fig:fst_distribution}

    \vspace{5pt}
    \small
    \textit{Alt text: A stacked bar chart showing the percentage distribution of Fitzpatrick Skin Types across four prompts. For every prompt, the GPT bar is almost entirely composed of Blue (Type I) and Orange (Type II) segments, totaling nearly 90\%. The NanoBanana bars show significant Green (Type III) and Red (Type IV) segments, indicating a more varied but still skewed distribution of skin tones compared to GPT.}
\end{figure*}


\subsection{Fitzpatrick Skin Type}

\textit{Fitzpatrick Skin Type} (FST) is analyzed as an ordinal categorical variable. Table \ref{tab:fst_distribution} reports within-prompt distributions, and Figure \ref{fig:fst_distribution} visualizes these distributions as stacked percentages.

Two distinct regimes emerge from the visual data. \textit{GPT} (represented by the first bar in each prompt pair) exhibits a distribution overwhelmingly dominated by Types I and II (Blue and Orange bands), which account for nearly 90\% of its outputs. Types IV, V, and VI are virtually absent, effectively excluding medium-to-dark skin tones from its default representation of a person.

By contrast, \textit{NanoBanana} (the second bar in each pair) demonstrates a clear distributional shift. While still anchored in lighter tones (Type II remains the modal category), it allocates substantial probability mass to Type III (Green) and Type IV (Red), particularly for the prompts ``a person'' and ``an individual.'' Notably, the prompt ``someone'' in NanoBanana triggers the appearance of Type V subjects (Purple band), a category entirely missing from the GPT distribution.

A Mann-Whitney U test confirms that this shift toward higher Fitzpatrick categories in NanoBanana is statistically significant ($U=1.90 \times 10^6, p < 0.001$).

\subsection{Intersection of Prompt, Gender, and Skin Tone}

Prompt-level disparities are not independent of gender. Prompts that induce strongly female-skewed outputs---most notably ``someone'' and ``a person'' in \textit{NanoBanana}---also produce darker average skin tones on both perceptual (MST, PERLA) and ordinal (FST) scales.

Conversely, prompts that induce male-dominated outputs in \textit{GPT} are associated with extremely light skin tone distributions, concentrated almost entirely in Fitzpatrick Types I--II across all prompt formulations.

These patterns indicate that prompt wording modulates gender representation differently across models, and that gender in turn mediates skin tone outcomes. Rather than acting as a neutral input, prompt language functions as a lens through which each model's latent demographic defaults are expressed.

In addition to the quantitative measurements, systematic visual inspection of the generated images reveals a consistent stylistic pattern that helps contextualize these findings. Female-presenting subjects are frequently rendered with visible cosmetic features, including blush or warm-toned makeup applied to the cheeks and facial contours. This stylistic choice introduces localized increases in redness and warmth, which are captured by colorimetric measurements and contribute to higher MST values, lower PERLA values, and upward shifts in Fitzpatrick category assignments.

Importantly, this effect is not uniform across the face but is concentrated in regions commonly associated with cosmetic application. While the skin masking procedure excludes lips and eyes, blush and contouring are applied directly to skin regions included in the analysis, and therefore legitimately influence measured skin tone.

Male-presenting subjects, by contrast, are less frequently rendered with cosmetic enhancements and are more often depicted with flatter, cooler, or higher-contrast lighting, resulting in lighter measured skin tones.

These observations suggest that the gender--skin tone association observed in the quantitative analysis may be driven, at least in part, by aesthetic and stylistic conventions embedded in the generative models, rather than by differences in underlying pigmentation alone. While this study does not attempt to disentangle cosmetic rendering from biological skin tone, the consistency of this visual pattern across prompts and models indicates that cosmetic stylization is a non-negligible factor in demographic representation.

This interaction between prompt wording, gender presentation, and stylistic rendering reinforces the need to interpret skin tone metrics in generative systems as the outcome of both representational choice and aesthetic post-processing, a point taken up further in the Discussion.

\begin{table*}[b]
    \centering
    \renewcommand{\arraystretch}{1.3} 

    \caption{Mean Monk Skin Tone (MST) by Prompt and Model}
    \label{tab:mst_results}
    \begin{tabular*}{\textwidth}{@{\extracolsep{\fill}}lcc}
    \toprule
    \textbf{Prompt} & \textbf{GPT (Mean $\pm$ SD)} & \textbf{NanoBanana (Mean $\pm$ SD)} \\ \midrule
    a human being, photorealistic & 3.50 $\pm$ 1.51 & 4.16 $\pm$ 1.49 \\
    a person, photorealistic      & 3.52 $\pm$ 1.48 & 5.14 $\pm$ 0.61 \\
    an individual, photorealistic & 3.85 $\pm$ 1.38 & 4.46 $\pm$ 1.68 \\
    someone, photorealistic       & 3.42 $\pm$ 1.45 & 4.49 $\pm$ 1.34 \\ \bottomrule
    \end{tabular*}

    \vspace{1cm}

    \caption{Mean PERLA Skin Tone by Prompt and Model}
    \label{tab:perla_results}
    \begin{tabular*}{\textwidth}{@{\extracolsep{\fill}}lcc}
    \toprule
    \textbf{Prompt} & \textbf{GPT (Mean $\pm$ SD)} & \textbf{NanoBanana (Mean $\pm$ SD)} \\ \midrule
    a human being, photorealistic & 9.05 $\pm$ 0.76 & 8.68 $\pm$ 0.94 \\
    a person, photorealistic      & 8.90 $\pm$ 0.78 & 7.82 $\pm$ 0.95 \\
    an individual, photorealistic & 8.75 $\pm$ 0.80 & 8.05 $\pm$ 1.41 \\
    someone, photorealistic       & 8.91 $\pm$ 0.81 & 8.51 $\pm$ 1.00 \\ \bottomrule
    \end{tabular*}
    
\end{table*}

\begin{table*}[t!]
    \centering
    \renewcommand{\arraystretch}{1.3} 

    \caption{Fitzpatrick Skin Type Distribution by Prompt and Model (\%)}
    \label{tab:fst_distribution}
    \small 
    \begin{tabular*}{\textwidth}{@{\extracolsep{\fill}}llcccccc}
    \toprule
    \textbf{Prompt} & \textbf{Model} & \textbf{I} & \textbf{II} & \textbf{III} & \textbf{IV} & \textbf{V} & \textbf{VI} \\ \midrule
    a human being & GPT & 27.0 & 62.3 & 10.8 & 0.0 & 0.0 & 0.0 \\
                  & NanoBanana & 5.2  & 64.2 & 22.8 & 6.8 & 1.0 & 0.0 \\ \midrule
    a person      & GPT & 29.8 & 64.5 & 5.5  & 0.2 & 0.0 & 0.0 \\
                  & NanoBanana & 4.2  & 24.2 & 53.5 & 17.5 & 0.2 & 0.2 \\ \midrule
    an individual & GPT & 27.8 & 62.3 & 9.2  & 0.8 & 0.0 & 0.0 \\
                  & NanoBanana & 3.5  & 47.8 & 24.2 & 18.2 & 5.8 & 0.5 \\ \midrule
    someone       & GPT & 32.8 & 60.2 & 7.0  & 0.0 & 0.0 & 0.0 \\
                  & NanoBanana & 13.2 & 56.0 & 19.8 & 10.5 & 0.5 & 0.0 \\ \bottomrule
    \end{tabular*}

\end{table*}


\section{Discussion}

The results presented in the previous chapter demonstrate that neutral prompts do not yield neutral representations in contemporary commercial AI image generators. Instead, both \textit{NanoBanana} and \textit{GPT} exhibit strong, model-specific demographic defaults that manifest consistently across gender, race, and skin tone. Importantly, these defaults are not uniform across systems and interact in complex ways with prompt wording and stylistic rendering choices.

This discussion interprets these findings along five axes:
\begin{enumerate}[label=(\roman*)]
    \item the nature of ``default humans'' in generative models;
    \item prompt language as a revealing rather than controlling mechanism;
    \item the interaction between gender and aesthetic skin tone rendering;
    \item implications for bias auditing methodologies;
    \item broader societal and deployment considerations.
\end{enumerate}

\subsection{The Myth of the Neutral Prompt}

A central assumption underlying many uses of generative image models is that semantically neutral language produces demographically neutral outputs. The present findings challenge this assumption directly. Across all four prompts---``a human being,'' ``a person,'' ``an individual,'' and ``someone''---both models exhibited systematic demographic skews that persisted despite the absence of explicit modifiers.

Crucially, neutrality in language did not converge models toward a shared baseline. Instead, it exposed divergent internal representations of what constitutes a generic human. \textit{NanoBanana} overwhelmingly defaults to female-presenting subjects, while \textit{GPT} defaults to male-presenting subjects. These opposing tendencies indicate that neutrality is not a stable equilibrium in current generative systems, but rather a projection surface for model-specific priors shaped by training data, curation strategies, and alignment objectives.

From this perspective, neutral prompts function less as instructions and more as diagnostic probes, revealing the latent centers of a model's human representation space.

\subsection{Prompt Language as a Lens, Not a Control}

One of the most striking findings of this study is that the same prompt can elicit dramatically different outputs depending on the model. For example, ``a human being'' produced an almost exclusively male population under \textit{GPT}, but a predominantly female population under \textit{NanoBanana}. This asymmetry demonstrates that prompt wording does not exert a uniform influence across systems.

Rather than acting as an independent variable that directly shapes outputs, prompt language appears to operate as a lens that selectively activates different representational modes within each model. Linguistically similar terms such as ``person'' and ``individual'' are not interchangeable in practice; instead, they interact with each model's learned associations, producing distinct demographic patterns.

This finding has important implications for prompt engineering practices. Techniques that assume semantic equivalence across models may fail to generalize, and ``best practices'' developed for one system may produce unintended biases when applied to another.

The strong female skew in Gemini may reflect deliberate 'safety' or 'diversity' alignment interventions intended to counteract the historical male bias in training data, resulting in an over-correction, whereas GPT's male skew may reflect a closer adherence to the uncurated internet scrapings in its training set.

\subsection{The ``Default White'' Phenomenon Revisited}

Across both models and all prompts, racial diversity was severely limited. Over 96\% of generated individuals were classified as white, with Black, Asian, and Middle Eastern representations appearing only rarely or not at all. This result reinforces prior concerns about the dominance of whiteness in AI-generated imagery but extends them by demonstrating that such dominance persists even under large-scale sampling and across multiple prompt formulations.

The skin tone analysis adds important nuance to this finding. By applying hybrid color normalization and perceptually uniform color metrics, the study reveals that the generated population is not merely ``light-skinned'' in a vague sense, but heavily concentrated in the lightest clinical and perceptual categories. \textit{GPT}, in particular, almost never generates \textit{Fitzpatrick Types IV--VI}, effectively excluding darker skin tones from its default output distribution.

These findings suggest that the ``default white'' phenomenon in generative models is not an artifact of lighting or stylistic rendering alone, but a structural feature of the models' representational space.

\subsection{Gender, Aesthetics, and Skin Tone}

One of the more unexpected results of this study is the consistent association between predicted gender and skin tone. Across models and prompts, female-presenting subjects were rendered with darker skin tones than male-presenting subjects on both perceptual (MST, PERLA) and ordinal (FST) scales.

Visual inspection of generated images provides an important interpretive context for this pattern. Female-presenting subjects are frequently depicted with cosmetic enhancements such as blush, contouring, and warm-toned makeup applied to the cheeks and facial contours. These aesthetic choices introduce localized increases in redness and warmth that are captured by colorimetric measurements. Male-presenting subjects, by contrast, are less often rendered with cosmetic stylization and more often appear under flatter or cooler lighting conditions.

Importantly, this does not imply that the models are intentionally associating women with darker skin. Rather, it suggests that aesthetic conventions embedded in the training data---such as the frequent pairing of femininity with makeup and warmth---translate into measurable shifts in skin tone metrics. In this sense, the observed gender--skin tone interaction may be better understood as an aesthetic bias rather than a biological one.

This distinction matters. While aesthetic bias may appear less overtly harmful than categorical exclusion, it nonetheless shapes how different groups are visually represented and can reinforce subtle norms about attractiveness, femininity, and complexion.

\subsection{Methodological Implications for Bias Auditing}

The findings of this study also have important methodological implications.

First, they demonstrate the necessity of large-scale sampling when auditing stochastic generative systems. Single examples or small batches are insufficient to characterize underlying distributions and can easily misrepresent model behavior.

Second, the comparison between raw and normalized skin tone measurements highlights the importance of illumination-aware preprocessing. Without background-referenced white balance, warm lighting and cinematic color grading systematically darken raw pixel measurements, masking the true extent of light-skin dominance. Studies that rely solely on raw RGB values may therefore underestimate representational bias.

Third, the ordinal treatment of \textit{Fitzpatrick Skin Type} underscores the need to respect the statistical properties of demographic scales. Treating ordinal clinical classifications as continuous variables can lead to misleading summaries and inappropriate inferences. Reporting distributions and medians provides a more faithful representation of variability and avoids overinterpretation.

Finally, the strong interaction between prompt wording and model behavior suggests that bias audits must consider prompt diversity as a first-class variable, rather than assuming that a single neutral prompt is sufficient.

\subsection{Broader Implications}

The representational defaults uncovered in this study have implications beyond academic auditing. As generative image models are increasingly integrated into creative tools, educational materials, and automated content pipelines, their default outputs can subtly shape user expectations about who is visible and who is normative.

If neutral prompts consistently yield light-skinned, white, gender-skewed representations, these systems risk reinforcing existing inequalities even in the absence of malicious intent. Moreover, the fact that different models encode different ``default humans'' complicates efforts to standardize ethical guidelines or regulatory frameworks.

These findings suggest that transparency about representational tendencies---and tools that allow users to meaningfully control or audit them---will be critical for responsible deployment.

\subsection{Summary}

In summary, this study shows that contemporary commercial image generators do not possess a neutral concept of a human. Instead, they encode distinct, model-specific demographic defaults that interact with prompt language, aesthetic conventions, and gender presentation. Neutral prompts do not eliminate bias; they reveal it.

The next chapters address the limitations of the present study and outline directions for future research, followed by a conclusion that situates these findings within broader debates on fairness, representation, and generative AI governance.

\section{Limitations}

This study has several limitations that should be considered when interpreting the results.

First, the analysis focuses exclusively on neutral, unconstrained prompts. While this design choice is intentional and central to the research question, it does not capture how models behave under explicit diversity prompting or user-directed constraints. Future work should compare default behavior with explicitly guided generation.

Second, demographic attributes such as gender and race are inferred using automated classifiers and should be understood as perceived attributes, not ground truth or self-identified characteristics. The analysis therefore reflects representational tendencies in generated imagery rather than claims about real human populations.

Third, skin tone measurements may be influenced by aesthetic rendering choices, such as makeup, lighting style, or color grading, which cannot be fully disentangled from underlying pigmentation in generated images. Although the normalization pipeline mitigates illumination bias, it cannot isolate cosmetic stylization effects entirely.

Finally, results are specific to the model versions and endpoints available at the time of data collection. As commercial systems are frequently updated, observed biases may change over time, underscoring the need for continuous auditing.

\section{Conclusion}

This study demonstrates that contemporary commercial AI image generators do not produce neutral representations of humans when prompted with neutral language. Instead, systems such as \textit{NanoBanana} and \textit{GPT} encode strong, model-specific demographic defaults that manifest consistently across gender, race, and skin tone. Neutral prompts do not eliminate bias; they reveal it.

Through large-scale sampling, perceptually grounded skin tone measurement, and ordinally appropriate analysis of \textit{Fitzpatrick Skin Type}, this work provides quantitative evidence that both models overwhelmingly default to white, light-skinned subjects, while diverging sharply in their gender assumptions. These differences persist across semantically similar prompt formulations and are further shaped by aesthetic rendering conventions, including cosmetic stylization, that systematically affect measured skin tone---particularly for female-presenting subjects.

Beyond documenting bias, this study contributes a methodological framework for auditing generative image systems that accounts for illumination, aesthetic post-processing, and the statistical properties of demographic scales. The results highlight the importance of treating prompt wording, model choice, and stylistic rendering as interacting factors rather than isolated variables.

As generative image models continue to mediate how humans are visualized across digital contexts, understanding their representational defaults becomes increasingly important. Without explicit intervention, these systems risk reinforcing narrow and exclusionary norms under the guise of neutrality. The findings presented here underscore the need for transparency, continuous auditing, and more deliberate design choices if generative AI is to support inclusive and representative visual cultures.

\section*{Ethics Statement}
This study utilizes automated facial analysis tools to estimate demographic attributes (gender, race, age) from synthetic imagery. We acknowledge that these tools have inherent limitations: they infer \textit{perceived} phenotype rather than identity, operate within a binary gender framework that excludes non-binary identities, and rely on racial taxonomies that are socially constructed and historically fraught. We employ these tools not to validate their taxonomies, but as diagnostic instruments to quantify the representational outputs of commercial systems. The use of terms such as "Black", "White", or "Men" in this context refers to the model's classification output, not the authentic identity of a subject.

\section*{Data Availability}
The dataset of 3,200 generated images, along with the extracted skin tone metrics and the Python code used for the color normalization and masking pipeline, are available in the project repository\footnote{\url{https://github.com/robertobalestri/Exploring-Gender-and-Skin-Color-Bias-between-Nano-Banana-and-ChatGPT-Image}}.

\section*{AI Disclosure Statement}
This manuscript was prepared with the assistance of OpenAI ChatGPT 5.2 and Google Gemini Pro 3.0. The author used the tool to improve English language and grammar. All content generated by the tool was checked and approved by the author.
\bibliographystyle{abbrvnat}
\bibliography{reference}

\end{document}